\documentclass[runningheads]{llncs}
\usepackage[T1]{fontenc}
\usepackage{float}
\usepackage{graphicx}
\usepackage{amsmath}
\usepackage{algorithm}
\usepackage{algpseudocode}
\usepackage[section]{placeins}
\begin{document}
\title{Single MCMC Chain Parallelisation on Decision Trees }
\author{Efthyvoulos Drousiotis\inst{1}\and
Paul G. Spirakis \inst{2,3}}
\authorrunning{E.Drousiotis et al.}
\institute{Department of Electrical Engineering and Electronics, University of %Liverpool, Liverpool L69 3GJ, UK;
\email{E.Drousiotis@liverpool.ac.uk} \and
Department of Computer Science, University of Liverpool, Liverpool L69 3BX, UK\\
\email{spirakis@liverpool.ac.uk}\and Department of Computer Engineering and Informatics, University of Patras, 26504 Patras, Greece}
\maketitle              % typeset the header of the contribution

\begin{abstract}
Decision trees are highly famous in machine learning and usually acquire state-of-the-art performance. Despite that, well-known variants like CART, ID3, random forest, and boosted trees miss a probabilistic version that encodes prior assumptions about tree structures and shares statistical strength between node parameters. Existing work on Bayesian decision trees depend on Markov Chain Monte Carlo (MCMC), which can be computationally slow, especially on high dimensional data and expensive proposals. In this study, we propose a method to parallelise a single MCMC decision tree chain on an average laptop or personal computer that enables us to reduce its run-time through multi-core processing while the results are statistically identical to conventional sequential implementation. We also calculate the theoretical and practical reduction in run time, which can be obtained utilising our method on multi-processor architectures. Experiments showed that we could achieve 18 times faster running time provided that the serial and the parallel implementation are statistically identical. 
\end{abstract}


\section{Introduction}\label{Section1}

In Bayesian statistics, it is a common problem to collect and compute random samples from a probability distribution. Markov Chain Monte Carlo (MCMC) is an intensive technique commonly used to address this problem when direct sampling is often arduous or impossible. MCMC using Bayesian inference is often used to solve problems in biology \cite{valderrama2019mcmc}, forensics \cite{taylor2014interpreting}, education \cite{drousiotis2021early}, and chemistry \cite{dumont2021quantification}, among other areas making it one of the most widely used algorithms when a collection of samples from a probability distribution is needed.
Monte Carlo applications are generally considered embarrassingly parallel since each chain can run independently on two or more independent machines or cores. Despite that, the main problem is that each chain is not embarrassingly parallel, and when the feature space and the proposal are computationally expensive, we can not do much to improve the running time and get  results faster. When we have to handle huge state-spaces and complex compound states, it takes significant time for an MCMC simulation to converge on an adequate model not only in terms of the number of iterations required but also the complexity of the calculations occurring in each iteration(such as searching for the best features and tree shape of a decision tree). For example, running an MCMC on a single chain Decision tree for a dataset of $400 000$ datapoints and $15$ features took upwards of 6 hours to converge when run on a $2.3 - 5.10 GHz$ Intel Core i7-10875H. In \cite{byrd2008speculative}, an approach aiming to parallelise a single chain is presented, and the improvement achieved is at its best 2.2 times faster. The functionality of this kind of solution is therefore limited as in real-time, and life applications run time is critical. The work presented in this paper aims to find methods to  significantly reduce the MCMC Decision tree's runtime by emphasising on the implementation of MCMC rather than the statistical algorithm itself. We aim to reduce significantly and up to an order of magnitude the run time of the MCMC Decision Tree on a single laptop or personal computer which is going to make the algorithm widely applicable and suitable for non tecinacl users. The remainder of this paper is organised as follows. Section 2 explains the MCMC in General and the Most Recent Work. Section 3 presents the MCMC in Decision trees. Our method is outlined in section 4, with the possible theoretical improvements. We introduce the case study in which we applied our method and reviewed results in section 5. Section 6 concludes the paper.

\section{Markov Chain Monte Carlo in General and Most Recent Work}\label{Section2}

One of the most widely used algorithms is the Metropolis \cite{metropolis1953equation} and its generalisation (see algorithm\ref{Metropolis Hashting}), the Metropolis-Hastings sampler (MH) \cite{hastings1970monte}. Given a partition of the state
vector into components, i.e., $x = (x_1, . . ., x_k )$, and that we wish to update the $i_th$ component, the Metropolis-Hastings update proceeds as follows. We first have a density for producing candidate observation $x'$, such that $x'_{i} = x_{i}$, which is denoted by $q(x,x')$. Given the chains ergodic condition, the definition of $q$ is arbitrary, and it has a stationary distribution $\pi$ which is selected so that the observations may be generated relatively easily.  After the new state generation $ x' = (x_1, . . ., x_{i-1},
x_i, x_{i+1}, . . ., x_k )$ from density $q(x,x')$, the new state is accepted or rejected using the Rejections Sampling principle with acceptance probability $\alpha(x,x')$ given by equation \ref{chain1}. If the proposed state is rejected, the chain remains in the current state.

It is worth mentioning that acceptance probability in this form is not unique, considering there are many acceptance functions that supplies a chain with the required properties. Nevertheless,  Peskun(1973) \cite{peskun1973optimum} proved that MH is the optimal one where the proper states are rejected least often, which maximises the statistical efficiency meaning that more samples are collected with fewer iterations.

\begin{equation}\label{chain1}
    a(\chi, \chi') =  \min(1, \frac{\pi(x')}{\pi(x')} \frac{q(\chi|\chi') }{q(\chi'|\chi) })
\end{equation}

% \begin{theorem}\label{ergodic}
 %For any ercodig Markov Chain, there is a unique steady-state probability vector $\overrightarrow{\pi}$ %that is the left eigenvector of p, such that $\eta(i,t)$ is the number of visits to state $i$ in $t$ %steps, then $\lim_{x\to\infty}\frac{\eta(i,t)}{t} = \pi(i)$, where $\pi(i)>0$ is the steady-state %probability or state $i$
%\end{theorem}

On a Markov process, the next step depends on the current state, which makes it hard for a single Markov chain to be processed contemporaneously by several processing elements.
Byrd \cite{byrd2008reducing}. proposed a method to parallelise a single Markov chain(Multithreading on SMP Architectures), where we consider backup move “B” in a separate thread of execution as it is not possible to determine whether move “A” will be accepted. If “A” is accepted, the backup move ‘B’ - whether accepted or rejected - must be discarded as it was based upon a now supplanted chain state. If “A” is rejected, control will pass to “B”, saving much of the real-time spent considering “A” had “A” and “B” been evaluated sequentially. Of course, we may have as many concurrent threads as desired. 

At this point, it is worth mentioning that the single chain parallelisation can become quickly problematic as the efficiency of the parallelisation is not guaranteed, especially for computationally cheap proposal distributions. Also, we need to consider that nowadays, computers make serial computations much faster than in 2008, when the single parallelisable chain was proposed.

Another way of making faster MCMC applications is to reduce the convergence rate by requiring fewer iterations. Metropolis-Coupled MCMC($(MC)^{3}$) utilised multiple MCMC \cite{altekar2004parallel} chains to run at the same time, while one chain is treated as the "cold" where its parameters are set to normal while the other chains are treated as "hot", which are expected to accept the proposed moves. The space will be explored faster through the "hot" chains than the "cold" as they are more possible to make disadvantageous transitions and not to remain at near-optimal solutions. The speedup increased when more chains and cores were added. 

Our work is focused on achieving a faster execution time of the MCMC algorithm on Decision trees through multiprocessor architectures. We aim to reduce the number of iterations while the number of samples collected is not affected. Multi-threading on SMP Architectures and $(MC)^{3}$ differs from our work as the former targets rejected moves as a place for optimisation, and the latter requires communication between the chains. Moreover, the aims are different as  $(MC)^{3}$ expands the combination of the chain, enhancing the possibilities of discovering different solutions and assisting avoid the simulation getting stuck in local optima.

\subsection{Probabilistic trees packages and level of parallelism}
Most of the existing probabilistic tree packages are only supported by the R programming language.

BART \cite{chipman2010bart} software included in the CRAN package \footnote{https://cran.r-project.org/web/packages/BART/index.html} supports multi threading based on OpenMP, where there are numerous exceptions for operating systems, so it is difficult to generalise. Generally, Microsoft Windows lacks OpenMP detection since the GNU autotools do not natively exist on this platform and Apple macOS since the standard Xcode toolkit is also not provided. The parallel package provides multi-threading via forking, only available in Unix. BART under CRAN uses parallelisation for the predict function and running concurrent chains.  

BartMachine \footnote{https://cran.r-project.org/web/packages/bartMachine/index.html}, which is written in Java and its interface is provided by rJava package, which requires Java Development Kit(JDK), provides multi-threading features similar to BART. BartMachine is recommended only for those users who have a firm grounding in the java language and its tools to upgrade the package and get the best performance out of it. Similar to BART, its parallelisation is based on running concurrent chains.

The rest of the available packages, BayesTree\footnote{https://cran.r-project.org/web/packages/BayesTree/BayesTree.pdf},dbarts\footnote{https://cran.r-project.org/web/packages/dbarts/index.html},Bartpy\footnote{https://pypi.org/project/bartpy/},XBART\footnote{https://jingyuhe.com/xbart.html} and imptree\footnote{https://cran.r-project.org/web/packages/imptree/index.html} does not support any kind of parallelisation.

Concurrent chains can not solve the problem of long hours of execution time. For example, if a single chain needs 50 hours to execute, 5 chains will still need 50 hours if run concurrently.  In contrast, in our case, a chain that serially needs 50 hours now takes approximately 2 hours for each chain. Moreover, we can to run concurrent chains where each chain is parallelised. If our implementation is compared to a package like BartMachine and BART, the runtime improvement we achieved is around 18 times faster, and if we compare it with a package that does not offer any parallelisation like most of the existing ones, the run time improvement for 5 chains is around 85 times faster.

\section{Markov Chain Monte Carlo in Decision Tree}\label{Section2}

A decision tree typically starts with a root node, which branches into possible outcomes. Each of those outcomes leads to additional decision nodes, which branch off into other possibilities ending up in leaf nodes. This gives it a treelike shape.

Our model describes the conditional distribution of $y$ given $x$, where $x$ is a vector of predictors $[x = (x_1,x_2,...,x_p)]$.
The main components of the $tree(T)$ includes the depth of the tree$(d(T))$, the features$(k(T))$ and the thresholds$(c(T))$ for each node where $k(T),c(T) \in \theta$, and the possibilities $p(Y|T,\theta,x)$ for each leaf node$(L(T))$.
If $x$ lies in the region
corresponding to the $i_th$ terminal node, then $y|x$ has distribution $f(y|\theta_i)$, where f represents a parametric family
indexed by $\theta_i$. The model is called a probabilistic classification tree, according to the quantitative response y.

As Decision Trees are identified by $(\theta, T)$, a
Bayesian analysis of the problem proceeds by specifying
a prior probability distribution $p(\theta ,T)$. Because $\theta$ indexes the parametric model for each $T$, it will usually be convenient to use the relationship

\begin{equation}\label{fullformula}
 p(Y_1:_N,T,\theta|x_1:_N) = p(Y|T,\theta,x)p(\theta|T)p(T)
\end{equation}

In our case it is possible to analytically obtain eq \ref{fullformula} and calculate the posterior of $T$ as follows:

\begin{equation}\label{labels probabiliteis}
     p(Y|T,\theta,x) = \prod_{i = 1}^{N} p(Y_i|x_i,T,\theta)   
\end{equation}

\begin{equation}\label{features and thresholds}
    p(\theta|T) = \prod_{j\in(T)}p(\theta_j|T) = \prod_{j\in(T)} p(k_j|T)p(c_j|k_j,T)
\end{equation}

\begin{equation}\label{prior}
    p(T) = \frac{a}{(1+d)^\beta}
\end{equation}

Equation \ref{labels probabiliteis} describes the product of the probabilities of every data point($Y_i$) classified correctly given the datapoints features($x_i$), the tree structure($T$), and the features/thresholds($\theta$) on each node on the tree.
Equation \ref{features and thresholds} describes the product of possibilities of picking the specific feature($k$) and threshold($c$) on every node given the tree structure($T$).
Equation \ref{prior} is used as the prior for tree $T_i$. This formula is recommended by \cite{chipman2010bart} and three aspects specify it: the probability that a node at depth $d (=0.1.2. . . .)$ is nonterminal, the parameter $a\in {0,1}$ which controls how likely a node would split, with larger $\alpha$ values increasing the probability of split, and the parameter ${\beta > 0}$ which controls the number of terminal nodes, with larger values of $\beta$ reducing the number of terminal nodes. This feature is crucial as this is the penalizing feature of our probabilistic tree which prevents it from overfitting and allowing convergence to the target function $f(X)$ \cite{rovckova2019theory}, and it puts higher probability on "bushy" trees, those whose terminal nodes do not vary too much in depth.

An exhaustive evaluation of equation \ref{fullformula} over all trees $T$ will not be feasible, except in trivially small problems, because of the sheer
number of possible trees, which makes it nearly impossible to determine precisely which trees have the largest posterior probability.

Despite these limitations, Metropolis-Hastings algorithms can still be used to explore the posterior. Such algorithms simulate a Markov chain sequence of trees such as:

\begin{equation}\label{chain}
    T_0, T_1, T_2,....,T_n
\end{equation}

which are converging in distribution to the posterior
$p(Y|T,\theta,x)p(\theta|T)p(T)$ in equtaion \ref{fullformula}.

Because such a simulated sequence will tend to gravitate toward regions of higher posterior probability, the simulation can be used to search for high-posterior probability trees stochastically. We next describe the details of such algorithms and their implementation.

\subsection{Specification of the Metropolis-Hastings
Search Algorithm on Decision Trees}

The Metropolis-Hastings(MH) algorithm for simulating
the Markov chain in Decision trees (see equation \ref{chain}) is defined as follows. Starting with an initial tree $T_0$, iteratively simulate the transitions from $T_i$ to $T_i+1$ by these two steps:

\begin{enumerate}
  \item Generate a candidate value $T'$ with probability distribution $q(T_i, T')$.
  \item Set $T_{i+1} = T'$ with probability
  \begin{equation}\label{chain}
    a(T_i, T') =  \min(1, \frac{\pi(Y_1:_N,T',\theta'|x_1:_N)}{\pi(Y_1:_N,T,\theta|x_1:_N)} \frac{q(T,\theta|T',\theta') }{q(T',\theta'|T,\theta) })
\end{equation}
Otherwise set $T_{i+1} = T_i$.
\end{enumerate}

To implement the algorithm, we need to specify the transition kernel $q$. We consider kernels $q(T, T')$, which generate $T'$ from $T$ by randomly choosing among four steps:

\begin{itemize}
\item Grow(G) : add a new $D(T)$ and choose uniformly a $k(T)$ and a $c(T)$
\item Prune(P) : choose uniformly a $D(T)$ to become a leaf
\item Change(C) = choose uniformly a $D(T)$ and change randomly a $k(T)$ and a $c(T)$
\item Swap(S) = choose uniformly two $D(T)$ and swap their $k(T)$ and $c(T)$
\end{itemize}

The rules are chosen by picking a number uniformly between 0 and 1 and each action have its own interval. For example, $p(G)=0.3 ,p(P)=0.3 ,p(C)=0.2 ,p(S)=0.2, [0, 0.3, 0.6, 0.8, 1]$

The probabilities (see equation \ref{Actions}) represent the sum of the probabilities of every accepted forward move. P(G), p(P), p(C), p(S) are set by the user who chooses how often each move wants to be proposed.

\begin{equation}\label{Actions}
q(T',\theta'|T,\theta) = q(T'|T)q(\theta'|T')=\sum_a q(a) q(T'|T,a)q(\theta'| T',\theta,a)
\end{equation}

where : 

\begin{equation}\label{Grow}
q(G)q(T'|T,G) q(\theta'|T',\theta,G) = p(G) \times \frac{1}{c} \times \frac{1}{k} \times \frac{1}{|L(T)|} 
\end{equation}

\begin{equation}\label{Prune}
q(P)q(T'|T,P) q(\theta'|T',\theta,P) =p(P) \times  \frac{1}{|D(T)| - 1} 
\end{equation}

\begin{equation}\label{Change}
q(C)q(T'|T,C) q(\theta'|T',\theta,c) = p(C) \times \frac{1}{ |D(T)|}\times \frac{1}{c} \times \frac{1}{k} 
\end{equation}

\begin{equation}\label{Swap}
q(S)q(T'|T,S) q(\theta'|T',\theta,S) =p(S) \times \frac{1}{ (|D(T)|(|D(T)| - 1)) / 2}    
\end{equation}

Equation \ref{Grow} can be described as the possibility of proposing the grow move including the probability of choosing the specific feature($k$), threshold($c$) and leaf node($|L(T)|$) to grow. P(G) is multiplied by the number of features($k$), the unique number of datapoints($c$) and the number of leaf nodes($|L(T)|$). For example, given a dataset with 100 unique datapoints($c$), 5 features($k$), a tree structure($T$) with 7 leaf nodes($|L(T)|$) and a $p(G)=0.3$ eq\ref{Grow} will be $0.3\times \frac{1}{100}\times \frac{1}{5}\times \frac{1}{7}$

Equation \ref{Prune} is the possibility of proposing the prune move, where p(P) is multiplied by the number of decision nodes subtracting one($(|D(T)| - 1)$ we are not allowed to prune the root node). In practise, given a $p(P)=0.3$ and a tree structure($T$) with 7 decision nodes($|D(T)|$) eq\ref{Prune} will be $0.3\times \frac{1}{10-1}$

Equation \ref{Change} is the possibility of proposing the change move where p(C) is multiplied by the number of decision nodes($|D(T)|$), the number of features($k$), and the number of unique datapoints($c$). For example, given a dataset with with 100 unique datapoints($c$), 5 features($k$), a tree structure($T$) with 12 decision nodes($|D(T)|$) and a $p(G)=0.2$ eq\ref{Grow} will be $0.2\times \frac{1}{100}\times \frac{1}{5}\times \frac{1}{12}$

Equation \ref{Swap} is the possibility of proposing the swap move where p(S) is multiplied by the number of paired decision nodes($|D(T)|$).In practise, given a tree structure($T$) with 12 decision nodes($|D(T)|$) and a $p(S)=0.2$ eq\ref{Swap} will be $0.2\times \frac{1}{((12)(12-1))/2}$

\begin{theorem}\label{testenv-theorem}Transition kernel(see equation \ref{reverseActions}) yields a reversible Markov chain, as every step from $T$ to $T'$ has a counterpart that can move from $T'$ to $T$. 
\end{theorem}

\begin{equation}\label{reverseActions}
q(T',\theta'|T,\theta)
\end{equation}

\begin{proof}
Assume a Markov chain, starting from its unique invariant distribution $\pi$. Now, take into consideration that for every sample $T_0,T_1,...,T_n$ have the same joint probability mass function(p.m.f) as their time reversal $T_n,T_{n-1},...,T_0$, so as we can call the Markov chain reversible, as well as its invariant distribution $\pi$ is reversible. This can be explained as a simulation of a reversible chain that looks the same if it runs backward.

The first thing we have to look for is if the Markov chain starts at $\pi$, and it can be checked by equation\ref{reversiblejump}

\begin{equation}\label{reversiblejump}
\begin{split}
&P(T_{k} = i|T_{k+1}=j, T_{k+2} = i_{k+2}, ...., T_n=i_n )\\
&
= \frac{P(T_{k} = i,T_{k+1}=j, T_{k+2} = i_{k+2}, ...., T_n=i_n )}{P(T_{k+1}=j, T_{k+2} = i_{k+2}, ...., T_n=i_n )}\\
&
=\frac{\pi P_{ij}P{ji_{k+2}} .... P_{i_{n-1}i_n}}{\pi P{ji_{k+2}} .... P_{i_{n-1}i_n}}\\
&
=\frac{\pi P_{ij}}{\pi_j}
\end{split}
\end{equation}

Equation\ref{reversiblejump} is only dependent on $i$ and $j$ where this expression for reversibility must be the same as the forward transition probability $P(X=T_{k+1} = i|X=T_k = j) = P_{ji}$. If, both original and the reverse Markov chains have the same transition probabilities, then their p.f.m must be the same as well.
\end{proof}

An example for our probabilistic tree is the following:

Assume a tree structure($T$) with 5 leaf nodes($|L(T)|$) and 11 decision nodes($|D(T)|$) sampling from a given dataset with 4 features($c$) and 100 unique datapoints($k$) for each feature.

If for example the forward proposal($q(T',\theta'|T,\theta)$) = ($"change"$) with p(C) = 0.2, we end up with the following equation: 
$0.2\times \frac{1}{11}\times\frac{1}{4}\times\frac{1}{100}$.
At the same time the reverse proposal(going from the current position to the previous) ($q(T,\theta|T',\theta')$) equation looks exactly the same as the  forward proposal. Given the above practical example we have strengthen our proof which shows that ($q(T',\theta'|T,\theta)$) = ($q(T,\theta|T',\theta')$), which shows in practise the reverse transition kernel nature of our model.

\begin{algorithm}
\caption{The General Metropolis-Hashting Algorithm}\label{Metropolis Hashting}
\begin{algorithmic}
\State Initialize $X_0$
\For{$i = 1$ to $N$}
    \State sample $\chi'$ from $q(\chi'|\chi_{t-1})$ 
    \State Calculate  $\alpha(\chi_i,\chi_{t-1}) =  \min(1, \frac{\pi(x')}{\pi(x')} \frac{q(\chi|\chi') }{q(\chi'|\chi) })$  
    \State Draw $u$ from $u[0,1]$
    \If{$u < \alpha(\chi_i|\chi_{t-1})$ }
        \State $\chi_i = \chi'$
    \Else
        
        \State $\chi_i = \chi_{t-1}$
\EndIf
\EndFor
\end{algorithmic}
\end{algorithm}

\section{Parallelising a single Decision Tree MCMC Chain}

Given a Decision's Tree MCMC chain with $N$ iterations, we propose a method that utilises $C$ number of cores aiming to enhance the running time of a single chain by at least an order of magnitude. As stated in Section\ref{Section1} and Section\ref{Section2}, at each iteration, a new sample $\chi'$ is drawn from the proposal distribution. Our method requires sampling from $C$ number of cores, $S (C=S)$ number of samples in parallel. We then accept the sample with the greatest $a(T_i, T')$ and repeat the same method until the Markov Chain converges to a stationary distribution. %In MCMC we can theoretically prove the chain convergence through coupling and total variation distance(see theorem \ref{theorem}).% 
In our method, we check the Markov chain convergence when the F1-score fluctuates less than $\pm 3\%$ for at least 100 iterations. Once the Chain has converged, we proceed to the second phase of our method. We now keep producing samples using $C$ cores, but we can now collect more than one sample which satisfies  $a(T_i, T') >= u$. ($u$ is a random uniform number$[0,1]$), otherwise we collect $T_i$
From this point, we will propose new samples from the sample with the greatest $a(T_i, T')$ until we are happy with the number of samples collected. Using this method, we can collect the same number of samples and explore the feature space as effectively as the serial implementation, but 18 times faster using an average laptop or personal computer. 

Our algorithm reduces the number of iterations and explores the feature space faster as we use more cores. This provides us with a significant run time improvement up to 18 times faster when the feature space is big and the proposal is expensive. The following sections will evaluate the running time improvement and the quality of the samples produced.

\begin{algorithm}
\caption{Single Chain parallelisation on MCMC Decision Trees}\label{DecisionTreeMetropolisParallel}
\begin{algorithmic}
\State Initialize $T_0$
\For{$i = 1$ to $N$}
    \State sample $C$ number of $T'$ from $Q(T{^j}'|T{^j}_{t-1})$ 
    \State Calculate in parallel $\alpha(T{^j}_i|T{^j}_{t-1}) =  \min(1,\frac{p(Y_1:_N,T',\theta'|x_1:_N)}{p(Y_1:_N,T,\theta|x_1:} \frac{q(T,\theta|T',\theta') }{q(T',\theta'|T,\theta) })$ for each sample
    \State Store every sampled posterior $\alpha(T{^j}_t|T{^j}_{t-1})$ value
    \State Until converge $T'$ = $\max{\alpha(T{^j}_i|T{^j}_{t-1})}$
    \If{Markov Chain converged}
    \State Draw $u$ from $uniform[0,1]$
        \For {$j= 1$ to $j$} \algorithmiccomment{For loop run in parallel}
            \If{$\alpha(T{^j}_i|T{^j}_{t-1}) > u$}
            \State Collect Sample $T'$
            \State $T' = \max{\alpha(T{^j}_i|T{^j}_{t-1}})$
            \Else
            \State Collect Sample $T$
        \EndIf
        \EndFor

\EndIf
\EndFor
\end{algorithmic}
\end{algorithm}

\subsubsection{Theoretical gains}
\begin{figure}[htbp]
\centerline{\includegraphics[width=0.85\textwidth]{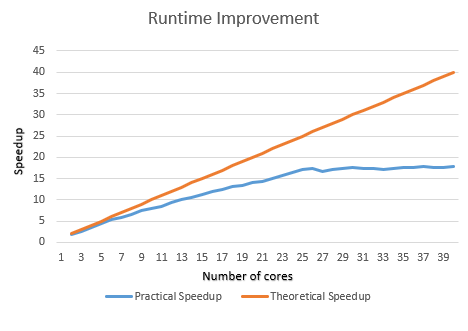}}
\caption{Speedup achieved by utilising different number of cores}
\label{speedupgraph}
\end{figure}

Using $C$ cores simultaneously, the programme cycle consists of repeated "steps," each performing the equivalent of between 1 and $n$ iterations. We need to calculate the number of iterations based on the acceptance rate to produce the same number of samples($S$) when we increase $C$. The moves are considered in parallel, where they are accepted or rejected. Given that the average probability of a single arbitrary move being rejected is $p_r$, the probability of the $i_{th}$ in every single concurrent core is $pr$. This step continues for $i$ iterations where equations \ref{iterations}, \ref{Runtime}, and \ref{speedup} show the iterations needed, run time, and speedup improvement, respectively, given a time($t$) in minutes for each iteration. Theoretical speedup (see figure\ref{speedupgraph}) were plotted for varying cores.

\begin{equation}\label{iterations}
i = \frac{S/C}{pr}
\end{equation}

\begin{equation}\label{Runtime}
Runtime = \frac{i}{t}
\end{equation}

\begin{equation}\label{speedup}
Speedup = \frac{Runtime}{C}
\end{equation}

For example, given a single MCMC Decision Tree chain running for $10000$ iterations and an acceptance rate of $70\%$,  after $3000$ burn-in iterations, we end up with $4900$ samples.

For the parallel MCMC chain, given the same settings as the serial one, with a $30\%$ burn-in period and 25 cores, we will collect the same number of samples with 500 iterations. This provides us with a 25 times faster execution time. Algorithm \ref{DecisionTreeMetropolisParallel} indicates the part implemented on a parallel environment, and figure 3 the maximum theoretical benefits from utilising our method.
Considering the communication overhead, the parts of the algorithm that are not parallelised, and the fact that the cores does not receive constant utilisation, in practise the speedups of this order are not achievable. Therefore, we will test it in practise and find out how it performs on real-life scenarios respect to the accuracy and the runtime improvement

\section{Results}

\subsection{Quality of the samples between serial and parallel implementation}

We have used the Wine dataset from scikit learn datasets\footnote{https://scikit-learn.org/stable/datasets/toydataset.html} repository as well as Pima Indians Diabetes and Dermatology from UCI machine learning repository \footnote{https://archive.ics.uci.edu/ml/index.php}, which are publicly available, to examine the quality of the samples on several testing hypotheses, including the different number of cores per iteration given the average F1-Score. Precision and Recall were also calculated for more depth and detailed insights about the performance and quality of the samples. The results, including the F1-Score, Precision, Recall, and Accuracy(see table2, table3 and table4) produced through 25-Fold Cross-Validation, ensure that every observation from the original dataset has the possibility of appearing in the training and test sets and also reduce any statistical error. Before any performance comparison, we need to examine whether the samples produced for each test case(25 cores and 40 cores) have any statistical difference from the serial implementation. We next examine if extracted samples by utilising 25 and 40 cores are representative of the family of the data they come. We use as ground truth data the F1-scores on each sample collected on every fold of the serial implementation, and we compare them with the corresponding collected samples from the other two test cases. In order to check any statistical difference between the samples, we performed the two-sample t-test for unpaired data\cite{fisher19783}, which is defined as follows: 

\begin{equation}\label{t_test}
T = \frac{Y_1 - Y_2}{\sqrt{\frac{s_{1}^{2}}{N_1}+\frac{s_{2}^{2}}{N_2}}}
\end{equation}

\begin{equation}\label{degrees of freedom}
|T| > t_{1-a/2,\nu}
\newline
where:
\newline
\nu =
\frac{
  \left(\dfrac{s_{1}^{2}}{n_1}+\dfrac{s_{2}^{2}}{n_2}\right)^{\!2}
}{
  \dfrac{(s_{1}^{2}/n^{}_1)^2}{n_1-1}+
  \dfrac{(s_{2}^{2}/n^{}_2)^2}{n_2-1}
}
\end{equation}

Formula\ref{t_test} is used to calculate the t-Test statistic equation where $N_1$ and $N_2$ are the sample sizes, $Y_1$ and $Y_2$ are the sample means, and $s_{1}^{2}$ and $s_{2}^{2}$ are the sample variances. The null hypothesis is rejected when equation \ref{degrees of freedom} holds, which is the critical value of the $t$ distribution with $\nu$ degrees of freedom.

For our first dataset(Wine), we first examine the serial implementation with the parallel using 25 cores. The absolute value of the t-Test, 0.62, is less than the critical value of 1.964, so we prove the null hypothesis and conclude that the samples drawn by using 25 cores have not any statistical difference at the 0.05 significance level.
We then compare the serial implementation with the parallel using 40 cores. In this case, the absolute value of the t-Test, 0.63, is less than the critical value of 1.964, so we prove the null hypothesis and conclude that the samples drawn by using 40 cores have not any statistical difference at the 0.05 significance level.
We also examine the parallel implementations between them(using 25 and 40 cores accordingly). In this case, the absolute value of t-Test, 0.64, is less than the critical value of 1.964, so we reject the null hypothesis and conclude that the samples drawn using 40 cores(in comparison with the samples drawn with 25 cores) have not a statistical difference at the 0.05 significance level.
The results for the rest of the datasets are presented explicitly in table 1

\begin{table}[htbp]\label{T}
\begin{tabular}{|l|c|c|c|c|}
\hline
Datasets & 1 vs 25 cores & 1 vs 40 cores & \multicolumn{1}{l|}{25 vs 40 cores} & \multicolumn{1}{l|}{Critical T value} \\ \hline
Pima Indians Diabetes & 0.51 & 0.63 & 0.63 & 1.97 \\
Dermatology & 0.73 & 0.77 & 0.77 & 1.97 \\
Wine & 0.62 & 0.64 & 0.65 & 1.97 \\ \hline
\end{tabular}\caption{Datasets Critical Values}
\end{table}

T-test proves that if we use up to 40 cores for sampling(rarely laptops and personal computers have more than 40cores), the quality of the samples are the same, ending up with statistically same samples as the serial implementation.Table 1, table 2, and table 3 shows that when we sample in parallel using up to 40 cores, the accuracy and the F1-score remain on the same levels as the serial implementation.
Tables 1, 2, 3, 4 indicate that a single chain on MCMC Decision Trees can not be an embarrassingly parallel algorithm as we can only improve the running time of a single chain by utilising a specific number of cores. The running time improvement we achieved($\times18$ faster) is the maximum run-time enhancement we can achieve on an MCMC decision tree to maintain the high metrics produced by the serial implementation. If we try to extract up to 40 samples per iteration, it is highly probable to get samples that does not affect the final results negatively. According to our results, the maximum number of cores can that be used is 40. Furthermore, Precision, Recall, and F1-score metrics indicate that no overfit is observed even when more than 3 labels exist, proving the samples' quality even in multi class classification problems.

\begin{table}[htbp]\label{tablewine}
\begin{tabular}{|l|cll|lll|lll|}
\hline
Labels & \multicolumn{3}{c|}{Precision} & \multicolumn{3}{c|}{Recall} & \multicolumn{3}{l|}{F1-score} \\ \hline
 & 1 core & 20cores & 40cores & \multicolumn{1}{c}{1 core} & 20cores & 40cores & \multicolumn{1}{c}{1 core} & 20cores & 40cores \\ \cline{2-10} 
0 & 0.85 & 0.88 & 0.88 & 1.00 & 1.00 & 1.00 & 0.92 & 0.94 & 0.94 \\
1 & 1.00 & 1.00 & 1.00 & 0.78 & 0.78 & 0.78 & 0.88 & 0.88 & 0.88 \\
2 & 1.00 & 0.93 & 0.93 & 1.00 & 1.00 & 1.00 & 1.00 & 0.96 & 0.96 \\ \hline
accuracy & \multicolumn{1}{l}{} &  &  &  &  &  & \textbf{0.93} & \textbf{0.93} & \textbf{0.93} \\ \hline
\end{tabular}\caption{Results for Wine Dataset}
\end{table}

\begin{table}[htbp]\label{tablepimaindiandiabetes}
\begin{tabular}{|l|cll|lll|lll|}
\hline
Labels & \multicolumn{3}{c|}{Precision} & \multicolumn{3}{c|}{Recall} & \multicolumn{3}{l|}{F1-score} \\ \hline
 & 1 core & 20cores & 40cores & \multicolumn{1}{c}{1 core} & 20cores & 40cores & \multicolumn{1}{c}{1 core} & 20cores & 40cores \\ \cline{2-10} 
0 & 0.83 & 0.84 & 0.84 & 0.86 & 0.83 & 0.83 & 0.84 & 0.84 & 0.84 \\
1 & 0.65 & 0.63 & 0.63 & 0.61 & 0.65 & 0.86 & 0.63 & 0.64 & 0.64 \\ \hline
accuracy & \multicolumn{1}{l}{} &  &  &  &  &  & \textbf{0.78} & \textbf{0.78} & \textbf{0.78} \\ \hline
\end{tabular}\caption{Results for Pima Indian Diabetes Datasets}
\end{table}

\begin{table}[htbp]\label{tabledermatology}
\begin{tabular}{|l|lll|lll|lll|}
\hline
Labels & \multicolumn{3}{c|}{Precision} & \multicolumn{3}{c|}{Recall} & \multicolumn{3}{l|}{F1-score} \\ \hline
 & \multicolumn{1}{c}{1 core} & 20cores & 40cores & \multicolumn{1}{c}{1 core} & 20cores & 40cores & \multicolumn{1}{c}{1 core} & 20cores & 40cores \\ \cline{2-10} 
0 & \multicolumn{1}{c}{0.93} & \multicolumn{1}{c}{0.93} & 0.93 & 1.00 & 1.00 & 1.00 & 0.97 & 0.97 & 0.97 \\
1 & 0.94 & 0.94 & 0.93 & 1.00 & 1.00 & 1.00 & 0.97 & 0.97 & 0.96 \\
2 & 1.00 & 1.00 & 1.00 & 0.96 & 0.96 & 0.96 & 0.98 & 0.98 & 1.00 \\
3 & 0.94 & 0.94 & 0.92 & 0.94 & 0.94 & 0.95 & 0.94 & 0.94 & 0.95 \\
4 & 1.00 & 1.00 & 1.00 & 1.00 & 1.00 & 1.00 & 1.00 & 1.00 & 1.00 \\
5 & 1.00 & 1.00 & 1.00 & 0.67 & 0.67 & 0.67 & 0.80 & 0.80 & 0.80 \\ \hline
accuracy &  &  &  &  &  &  & \textbf{0.96} & \textbf{0.96} & \textbf{0.96} \\ \hline
\end{tabular}\caption{Results for Dermatology Datasets}
\end{table}

%\begin{itemize}
%  \item Precision: the proportion of positive identifications which was actually %correct
%  \item Recall: the proportion of actual positives that were identified correctly
%  \item F-1 score: the weighted average of Precision and Recall columns
%  \item Accuracy: the ratio of correctly predicted observations over the total %observation
%\end{itemize}

%\begin{equation}
%Precision = \frac{TruePositives}{TruePositives+FalsePositives}\label{eq:3}
%\end{equation}
%\begin{equation}
%Recall =  \frac{TruePositives}{TruePositives+FalseNegatives}\label{eq:4}
%\end{equation}
%\begin{equation}
%F1 = 2\times \frac{Precision \times Recall}{Precision+Recall}\label{eq:5}
%\end{equation}

\subsection{Practical gains}

Figure 1 presents the theoretical and practical speedup achieved given the number of cores used demonstrating a remarkable runtime improvement, especially when the feature space is ample and the proposal is expensive. Figure 1 demonstrates that in practise, theoretical speedups of this order can not be achieved for various reasons, including communications overhead, and as well as the cores do not receive constant utilisation. The practical improvement achieved used the novel method we proposed, speeds up the process up to 18 times depending on the number of cores the user may choose to utilise. Moreover, figure 1 demonstrates that even if we use more than 25 cores, the speedup achieved is the same, because of the architecture of the cores. When we run on a local machine(laptop or personal computer) a medium size dataset(500,000 entries), the memory is not enough to run every single parallel tree on a separate core. Given that, we have to wait for a core to finish the task, in order to allocate its memory to another core. Given that, we end up that it is not always beneficial to use more cores, as faster execution time and speedup is not guaranteed. Scaling up to 25 cores is the ideal, having in mind that any number above that, might not benefit the run time.    
To the best of our knowledge it is the first time where a single chain in general, specifically on decision trees, is parallelised with our proposed method.

\section{Conclusion}

Our novel proposed method for parallelising a single MCMC Decision tree chain takes advantage of multicore machines without altering any properties of the Markov Chain. Moreover, our method can be easily and safely used in conjunction with other parallelisation strategies, i.e., where each parallel chain can be processed on a separate machine, each being sped up using our method.

Furthermore, our approach can be applied to any MCMC Decision tree algorithm which needs to process hundreds of thousands of data given an expensive proposal where an execution time of 18 times faster can be easily achieved. As multicore technology improves, CPUs with multiple processing cores will provide speed-ups closer to the theoretical limit calculated. By taking advantage of the improvements in modern processor designs our method can help make the use of MCMC Decision tree-based solutions more productive and increasingly applicable to a broader range of applications.
Future work includes on expanding our method on a High Performance Computer(HPC), servers, and cloud which are build for this kind of tasks to compare and demonstrate possible runtime improvements, and discover the merits of such technologies. Moreover, we plan to implement more MCMC single chain parallelisation techniques, including data partitioning, and conduct experiments with various size and shapes datasets, to find the most effective technique, given the type and shape of the dataset.   

%%
%% Bibliography
%%

%% Please use bibtex, 

\bibliography{decisiontree}

\end{document}